\documentclass{article}
\usepackage{spconf,amsmath,graphicx,hyperref}

\usepackage{cite}
\usepackage{amsmath,amssymb,amsfonts}
\usepackage{algorithmic,multirow}
\usepackage{graphicx,verbatim}
\usepackage{textcomp}
\usepackage{xcolor}
\usepackage{soul}

\title{Learning Structural--Functional Brain Representations through Multi--Scale Adaptive Graph Attention for Cognitive Insight}
\name{Badhan Mazumder$^{\star \dagger}$ \quad Sir-Lord Wiafe$^{\star \dagger}$ \quad Aline Kotoski$^{\ddagger \dagger}$ \quad Vince D. Calhoun$^{\star \dagger}$ \quad Dong Hye Ye$^{\star \dagger}$}
  
  \address{$^{\star}$ Department of Computer Science, Georgia State University\\
  $^{\ddagger}$Neuroscience Institute, Georgia State University\\
      $^{\dagger}$Tri-Institutional Center for Translational Research in Neuroimaging and Data Science (TReNDS),\\ Georgia State University, Georgia Institute of Technology, and Emory University}
\makeatletter
\let\oldthebibliography\thebibliography
\def\thebibliography#1{%
  \oldthebibliography{#1}%
  \setlength{\itemsep}{0pt}%
  \setlength{\parskip}{0pt}%
  \setlength{\parsep}{0pt}%
}
\makeatother
\begin{document}
%
\maketitle

\begin{abstract}

Understanding how brain structure and function interact is key to explaining intelligence yet modeling them jointly is challenging as the structural and functional connectome capture complementary aspects of organization. We introduced \textbf{M}ulti-scale \textbf{A}daptive \textbf{G}raph \textbf{Net}work (\textit{MAGNet}), a Transformer-style graph neural network framework that adaptively learns structure-function interactions. \textit{MAGNet} leverages source-based morphometry from structural MRI to extract inter-regional morphological features and fuses them with functional network connectivity from resting-state fMRI. A hybrid graph integrates direct and indirect pathways, while local-global attention refines connectivity importance and a joint loss simultaneously enforces cross-modal coherence and optimizes the prediction objective end-to-end. On the ABCD dataset, \textit{MAGNet} outperformed relevant baselines, demonstrating effective multimodal integration for advancing our understanding of cognitive function.

\end{abstract}

\vspace{-10pt}
\section{Introduction}
\vspace{-10pt}

A core challenge in neuroscience is explaining how structural and functional connectivity jointly enable cognition. Among cognitive abilities, intelligence is a central, multi-dimensional construct encompassing fluid intelligence (problem-solving and logical reasoning in novel scenarios) \cite{base4,exp_8}, crystallized intelligence (applying learned knowledge to resolve problems) \cite{exp_8}, and total intelligence (overall cognitive performance) \cite{exp_8}. Because these dimensions arise from the interplay between anatomical architecture and dynamic activity \cite{I_1}, integrative analyses of structural and functional systems are essential.


Source-based morphometry (SBM) from structural MRI (sMRI) maps gray-matter covariation, while functional network connectivity (FNC) from resting-state fMRI (rs-fMRI) captures temporal co-activation. Extracted via independent component analysis (ICA), both reside in a shared representational space that enables coherent multimodal fusion. Prior studies \cite{I_1,I_4} showed that jointly modeling morphology-informed and functional features clarifies neural correlates of cognitive performance, motivating deeper study of their combined contribution to individual differences in intelligence.

Graph neural networks (GNNs) have emerged for structure-function analysis \cite{My_1,My_2,My_3,My_4}, but typical approaches \cite{base6,base7,base5} employed the structural connectome as a static adjacency, limiting dynamic learning of structure-function interplay. Other methods \cite{base6,base8} processed structure and function separately with late fusion, under-capturing cross-modal interactions. Many models \cite{base7,base6} further assumed uniform structure-function coupling and emphasize local aggregation \cite{base4,base5,base6,base7}, overlooking global dependencies essential for functional network organization and neglecting indirect structural pathways (Detours \cite{base3}) that mediated interactions between networks without direct links.

We addressed these gaps by constructing a hybrid brain graph that integrated SBM with FNC, incorporating unimodal, cross-modal, and multi-scale detour connections to capture both direct and indirect structure-function interactions. Building on this, we proposed the \textit{MAGNet}, a Transformer-style GNN framework that first applied local edge-aware attention to selectively aggregate structure-function relations, then performed global self-attention to encode long-range dependencies and improve structure-function alignment. To enforce biologically meaningful coupling, we introduced a joint loss that combined a structure-function consistency term with the downstream prediction objective.

\begin{figure*}[t]
\centerline{\includegraphics[width=14 cm,height=6.5 cm]{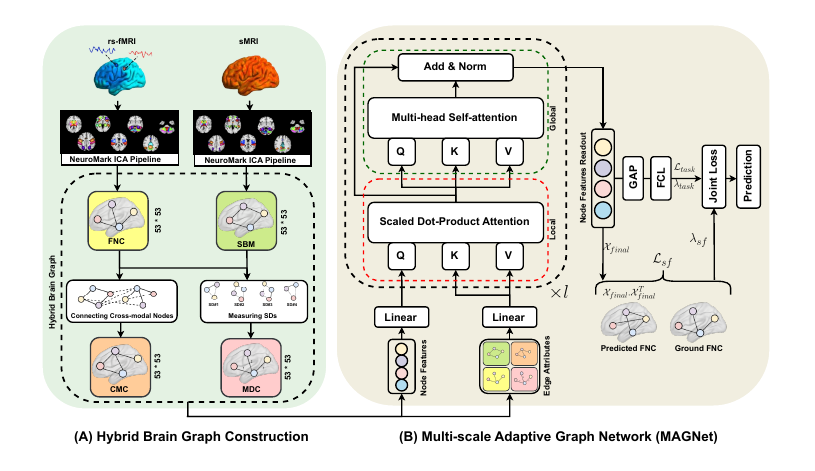}}
\vspace{-15pt}
\caption{(A) After generating FNC from rs-fMRI and SBM from sMRI, a hybrid brain graph was constructed with unimodal, cross-modal (CMC) and multi-scale detour connections (MDC) measuring structural detour (SD). (B) \textit{MAGNet}'s flowchart: scaled dot-product attention enabled local message passing with node features and edge attributes, followed by multi-head self-attention for global refinement for $l$ layers. Refined node embeddings then underwent global average pooling (GAP) and a fully connected layer (FCL) for intelligence score prediction optimized with a joint loss.} \label{fig1}
\vspace{-12 pt}
\end{figure*}

In summary, our major contributions are three-fold:
\begin{enumerate}
    \item We constructed a hybrid brain graph with unimodal, cross-modal, and multi-scale detour connections for comprehensive connectivity representations.
    \vspace{-5pt}
    \item We designed an adaptive GNN framework with local-global attention that prioritizes connectivity types, yielding neurologically grounded embeddings. 
    \vspace{-5pt}
    \item We formulated a joint objective combining structure-function consistency and prediction losses to improve alignment and end-to-end performance.
\end{enumerate}

\vspace{-10pt}
\section{Methodology}
\vspace{-10pt}

Assume $\mathcal{D}=\{((M_n^{s},M_n^{f}),G_n^{t})\}_{n=1}^{P}$, where $P$ is the number of subjects; for subject $n$, $M^{s}$ and $M^{f}$ denote structural and functional modalities, and $G^{t}$ with $t\in\{t_1,t_2,t_3\}=\{\text{fluid},\text{crystallized},\text{total}\}$ are the intelligence scores. As in Fig.~\ref{fig1}A, we build a hybrid brain graph linking $M^{s}=(O,A^{s},W^{s})$ and $M^{f}=(O,A^{f},W^{f})$, where $O=\{o_\epsilon\}_{\epsilon=1}^{\eta}$ (with $\eta$ brain networks) and edges $A^{s}=O\times O$, $A^{f}=O\times O$ are weighted by $W^{s},W^{f}\in\mathbb{R}^{\eta\times\eta}$. SBM is typically a group-wise $P\times\eta$ matrix (rows are $\eta$-dimensional loading vectors); we converted it to subject-specific SBM matrices $M_n^{s}\in\mathbb{R}^{\eta\times\eta}$ via vector transposition and multiplication, modeling gray matter volume (GMV) covariation to capture structure-function interaction. From $W^{s}$ and $W^{f}$, we formed k-nearest neighbor (k-NN) graphs by keeping the top $k$ edges per node to promote sparsity and relevance. Our objective is to obtain refined node embeddings that leverage structure-function interaction for improved intelligence score prediction at the graph level.

\vspace{-13 pt}
\subsection{ Hybrid Brain Graph Construction}
\vspace{-5pt}
The retained edges from $ W^{s} $ and $ W^{f} $ define the direct structural and functional pathways, which we considered as unimodal direct connections as : $ \textbf C^s = \{ C^s_{i,j} \ | \ (i,j) \in W^s \}, \quad \textbf C^f = \{ C^f_{i,j} \ | \ (i,j) \in W^f \}$ where $  \textbf C^s $ and $  \textbf C^f $ denote the sets of direct structural and functional connections, respectively, and each element $ C^s_{i,j} $ or $ C^f_{i,j} $ defines the connection between node (network) $i$ and $j$. 

To establish high-level associations between modalities, we constructed cross-modal connections (CMC) by quantifying the relationship between node connectivity across structural and functional modalities using cosine similarity as: $\textbf C^{cmc} = \{ C_{i,j}^{cmc} \ | \ C_{i,j}^{cmc} = \frac{E^s_i \cdot E^f_j}{| E^s_i | | E^f_j |}, \quad (i,j) \in W^s \times W^f \}$ where $E^s_i$ and $E^f_j \in \mathbb{R}^{\eta}$ represent the connectivity profiles of node $i$ in $W^s$ and node $j$ in $W^f$, respectively. The term $C_{i,j}^{cmc} \in \mathbb{R}$ captures the degree of alignment between the connectivity profiles of the two nodes across both modalities. Each node was linked to its top $\gamma (=8)$ most similar nodes from the other modality, ensuring effective cross-modal links.

Detour connectivity (DC) \cite{base3} quantifies alternative pathways within the structural connectivity by identifying structural detours (SD) defining indirect connections between functionally linked regions via Depth-First Search (DFS). However, standard DC relies on a fixed search radius $r$, limiting its ability to capture multi-scale structural interactions. In our work, we adapted and proposed multi-scale detour connectivity (MDC), which extends DC in SBM by incorporating short, medium, and long-range detour pathways, ensuring a hierarchical representation of indirect structural connections as follows: $\textbf C^{mdc} = \{ C^{mdc}_{i,j}(r) \ | \ (i,j) \in \bar{W}^s \times \bar{W}^f, \quad r \in \{r_1, r_2, r_3\} \}$, where $\bar{W}^{*}$ defines the binary adjacency matrix form and $ \textbf C^{mdc} $ represents the set of MDC matrices, each corresponding to a different search radius $ r \in \{r_1, r_2, r_3\} $, preserving connectivity for short ($r_1=2$ hop), medium ($r_2=3\quad\text{or}\quad 4$ hop), and long ($r_3\geq5$ hop) detours. Here, $ C^{mdc}_{i,j}(r)=|DFS(\bar{W}^{s}, i, j, r)|\quad \text{if} \quad \bar{W}^{f}_{i,j} > 0 ,\quad \text{otherwise} \quad 0$. The function $ DFS(\bar{W}^{s}, i, j, r) $ returns the number of valid detour paths between nodes $ i $ and $ j $ within the given radius $ r $. By capturing indirect interactions across multiple scales, MDC  offers a comprehensive view of structural dependencies shaping functional organization. The final hybrid graph $\rho$ integrates all four edge types, with node features as per-network averages of structure-function strengths.
\begin{equation}
\rho  = \textbf C^{s} \cup \textbf C^{f} \cup \textbf C^{cmc} \cup \textbf C^{mdc} 
\label{eq0_1}
\end{equation}

\vspace{-15 pt}
\subsection{Multi-scale Adaptive Graph Network (MAGNet)}
\vspace{-5 pt}
\subsubsection{Local Edge-Aware Message Passing}  
\vspace{-5 pt}
\textit{MAGNet} updates node features through an edge-aware attention mechanism, where information is selectively aggregated from neighboring nodes based on both node features and the type of connection between them. This mechanism enables the model to differentiate among edges in the hybrid brain graph \( \rho \) (Eq. \ref{eq0_1}), which includes structural, \( \mathbf{C}^s \), \( \mathbf{C}^f \), \( \mathbf{C}^{cmc} \), and multi-scale detour \( \mathbf{C}^{mdc} \) connections. For a node \( i \), the updated representation was computed by attending to its neighbors \( j \in \mathcal{N}(i) \subseteq \rho \), where each neighbor contributed proportionally based on an attention weight derived from both node features and edge attributes indicating the source modality of the connection. This allowed the model to learn attention patterns specific to the type of structure-function relationship. The node update is defined as:
\begin{equation}
\mathbf{z}_i = \sum_{j \in \mathcal{N}(i) \subseteq \rho} \frac{\exp \left( \frac{ \mathbf{Q}_i \cdot \mathbf{K}_{i,j}^{\rho} }{\sqrt{h}} \right)}{\sum_{j \in \mathcal{N}(i) \subseteq \rho} \exp \left( \frac{ \mathbf{Q}_i \cdot \mathbf{K}_{i,j}^{\rho} }{\sqrt{h}} \right)} \cdot \mathbf{V}_{i,j}^{\rho},
\label{eq:message_passing_rho}
\end{equation}
where \( \mathbf{Q}_i \) defines the query vector derived from the features of node \( i \), and \( \mathbf{K}_{i,j}^{\rho} \) and \( \mathbf{V}_{i,j}^{\rho} \) are the key and value vectors computed from the edge attributes \( \mathbf{e}_{i,j}^{\rho} \), which encode the connection type (e.g., FNC, SBM, CMC, MDC) from graph \( \rho \). \( h \) is the hidden dimension, and \( \mathcal{N}(i) \subseteq \rho \) denotes the neighborhood of node \( i \) defined by the edge structure of \( \rho \). 
\vspace{-10 pt}
\subsubsection{Global Self-Attention Refinement}  
\vspace{-5 pt}
After the initial edge-aware message passing, where each node aggregates information from its local neighbors to produce an updated feature representation $ \mathbf{z}_i $, the node features were assembled into a matrix form: $\mathcal{X} = [\mathbf{z}_1, \mathbf{z}_2, ..., \mathbf{z}_\eta]^\top \in \mathbb{R}^{\eta \times h}$; where each row of $ \mathcal{X} $ corresponds to the feature vector of a node after local attention-based aggregation. To further refine these node embeddings, we applied a global self-attention mechanism, allowing all nodes to interact and capture long-range dependencies within the graph. This was obtained employing a multi-head self-attention mechanism followed by a residual connection and layer normalization: $\mathcal{X}_{final} = LN(\mathcal{X} + MultiHeadAttention(\mathcal{X}, \mathcal{X}, \mathcal{X})),$ where $LN$ is layer normalization and $\mathcal{X} $ serves as the query, key, and value matrices, enabling global feature integration.

\vspace{-10 pt}
\subsubsection{Joint Loss Optimization}  
\vspace{-5 pt}
To train \textit{MAGNet}, we used a composite loss function combining a task-specific loss $\mathcal{L}_{task}$ and a structure-function consistency loss $\mathcal{L}_{sf}$. $\mathcal{L}_{task}$ measures the difference between the predicted graph-level intelligences scores $\hat{G}^{t}$ and the ground truth ${G^{t}}$ using mean squared error (MSE) as:$\mathcal{L}_{task} = \frac{1}{B} \sum_{b=1}^{B} ||G_b^{t} - \hat{G}_b^{t}||_2^2,$ where $B$ denotes the number of graphs per batch. Since structure and function are neurologically intertwined, we enforced that structure-function alignment on the refined node embeddings by generating predicted FNC matrix: $\hat{W^{f}} = \mathcal{X}_{final} \mathcal{X}_{final}^\top$ and measuring MSE with the ground-truth FNC matrix $W^{f}$ to define $\mathcal{L}_{sf}$ as follows: $\mathcal{L}_{sf} = \frac{1}{\eta ^2} \| \hat{W^{f}} - W^{f} \|_F^2$, where $\| \cdot \|_F^2$ defines the squared Frobenius norm. The proposed final joint loss term combined both objectives as follows: 
\begin{equation}
\mathcal{L}_{joint} = \lambda_{sf} \mathcal{L}_{sf} + \lambda_{task} \mathcal{L}_{task}   
\label{eq4}
\end{equation}
where $\lambda_{sf}$ and $\lambda_{task}$ balance the influence of the consistency term and the task term, respectively.

\vspace{-10 pt}
\section{Experiments and Results}
\vspace{-10 pt}
\subsection{Dataset and Preprocessing}
\vspace{-5 pt}

The Adolescent Brain Cognitive Development (ABCD) study is a nationwide longitudinal cohort across 21 U.S. sites examining determinants of physical and mental health from childhood through adolescence \cite{D_1}. We utilized baseline data from 7,656 children (Male: 3,951; Female: 3,705; ages 9–10) with rs-fMRI and sMRI, alongside fluid, crystallized, and total composite intelligence assessments.

Standard preprocessing (slice-timing correction, realignment, spatial normalization, smoothing) was applied to fMRI. The $NeuroMark$ \cite{D_2} ICA pipeline extracted $53$ intrinsic connectivity networks (ICNs). Signals were bandpass filtered ($0.01$–$0.15$ Hz), z-scored, and normalized dynamic time warping (nDTW)\cite{D_3} was used to form $53\times53$ FNC matrices. Following previous work \cite{D_3}, the nDTW window was set by the low-frequency limit yielded $110$ time points (TR = $0.8$ s for ABCD).

For sMRI, we performed fully automated constrained ICA with the $NeuroMark$ \cite{D_2} $53$-component template. SBM estimated subject-specific loadings, identifying spatially independent structural patterns across the $53$ areas \cite{I_4}. Using GMV as input, we obtained independent components capturing co-varying structural patterns, and encoded each subject’s structural features as a $53$-dimensional vector.

\vspace{-15 pt}
\subsection{Experimental Settings}
\vspace{-5 pt}
To identify the optimal hyperparameters for \textit{MAGNet}, we performed a grid search. Each configuration was assessed using 5-fold CV with an $80:20$ split ratio. The best-performing model utilized $4$ attention heads, a batch size of $16$, and a learning rate of $0.0001$. In all experiments, a hidden dimension of $64$, a dropout rate of $0.2$, and loss weightings of $\lambda_{sf} = 0.3$ and $\lambda_{task} = 0.7 $ were maintained while optimized using Adam for $50$ epochs. Predictive performance was evaluated by measuring mean square error (MSE), mean absolute error (MAE) and correlation, with results reported as the mean $\pm$ standard deviation over five experimental runs. 
For k-NN and MDC, we evaluated multiple settings and selected $k=5$ with $r \in \{2, 3, 5\}$ for this study.
\begin{table*}[t]
\caption{Comparison with SOTA baselines (mean$\pm$standard deviation).} \label{tab1}
\centering
\resizebox{\textwidth}{!}{
\begin{tabular}{|c|c|c|c|c|c|c|c|c|c|} 
\hline
 Method  & \multicolumn{3}{c|}{Fluid} & \multicolumn{3}{c|}{Crystallized} & \multicolumn{3}{c|}{Composite} \\  
\cline{2-10}
 & MSE & MAE & Correlation & MSE & MAE & Correlation & MSE & MAE & Correlation \\  \hline

 GAT \cite{base1}& 118.40$\pm$2.58 & 10.38$\pm$0.10 & 0.15$\pm$0.0145 & 62.57$\pm$3.31 & 7.63$\pm$0.17 & 0.11$\pm$0.0238 & 88.57$\pm$3.17 & 9.24$\pm$0.15 & 0.16$\pm$0.0238 \\  \hline

 GT \cite{base2}& 116.02$\pm$5.56 & 9.64$\pm$0.26 & 0.24$\pm$0.0150 & 57.15$\pm$5.36 & 6.90$\pm$0.36 & 0.26$\pm$0.0229 & 85.74$\pm$6.06 & 8.40$\pm$0.29 & 0.26$\pm$0.0187 \\ \hline

 SFDN  \cite{base3}& 112.43$\pm$0.80 & 9.15$\pm$0.41 & 0.29$\pm$0.0162 & 53.69$\pm$0.81 & 6.04$\pm$0.42 & 0.34$\pm$0.0135 & 83.46$\pm$0.81 & 8.02$\pm$0.41 & 0.35$\pm$0.0148 \\ \hline

 SFIN  \cite{base4}& 113.23$\pm$0.51 & 9.66$\pm$0.15 & 0.30$\pm$0.0273 & 54.91$\pm$0.32 & 6.16$\pm$0.27 & 0.32$\pm$0.0226 & 84.21$\pm$0.45 & 8.11$\pm$0.65 & 0.33$\pm$0.0221 \\ \hline

 Joint GCN \cite{base5}& 115.29$\pm$0.31 & 9.91$\pm$0.58 & 0.27$\pm$0.0186 & 55.81$\pm$0.53 & 6.32$\pm$0.50 & 0.28$\pm$0.0389 & 84.73$\pm$0.79 & 8.57$\pm$0.43 & 0.30$\pm$0.0263 \\ \hline

 BrainNN \cite{base6} & 113.60$\pm$1.59 & 9.06$\pm$0.13 & 0.28$\pm$0.0158 & 56.19$\pm$1.38 & 6.45$\pm$0.21 & 0.30$\pm$0.0161 & 86.65$\pm$1.29 & 8.19$\pm$0.16 & 0.32$\pm$0.0191 \\ \hline

 GCNN  \cite{base7}& 116.16$\pm$1.54 & 9.56$\pm$0.18 & 0.25$\pm$0.0151 & 58.10$\pm$1.05 & 6.95$\pm$0.12 & 0.29$\pm$0.0197 & 87.78$\pm$1.38 & 8.38$\pm$0.59 & 0.31$\pm$0.0177 \\ \hline

 Joint DCCA \cite{base8}& 116.90$\pm$1.83 & 9.21$\pm$0.86 & 0.26$\pm$0.0170 & 59.05$\pm$2.10 & 6.69$\pm$0.36 & 0.33$\pm$0.0232 & 84.97$\pm$2.34 & 8.60$\pm$0.51 & 0.34$\pm$0.0284 \\ \hline

 \textbf{\textit{MAGNet}} & \textbf{109.61$\pm$0.45} & \textbf{8.30$\pm$0.15} & \textbf{0.33$\pm$0.0137} & \textbf{50.47$\pm$0.21} & \textbf{5.53$\pm$0.24} & \textbf{0.36$\pm$0.0194} & \textbf{81.35$\pm$0.28} & \textbf{7.14$\pm$0.31} & \textbf{0.38$\pm$0.0184} \\ \hline

\end{tabular}
}\vspace{-15pt} 
\end{table*}
\begin{figure}[t]
\centerline{\includegraphics[width=\columnwidth]{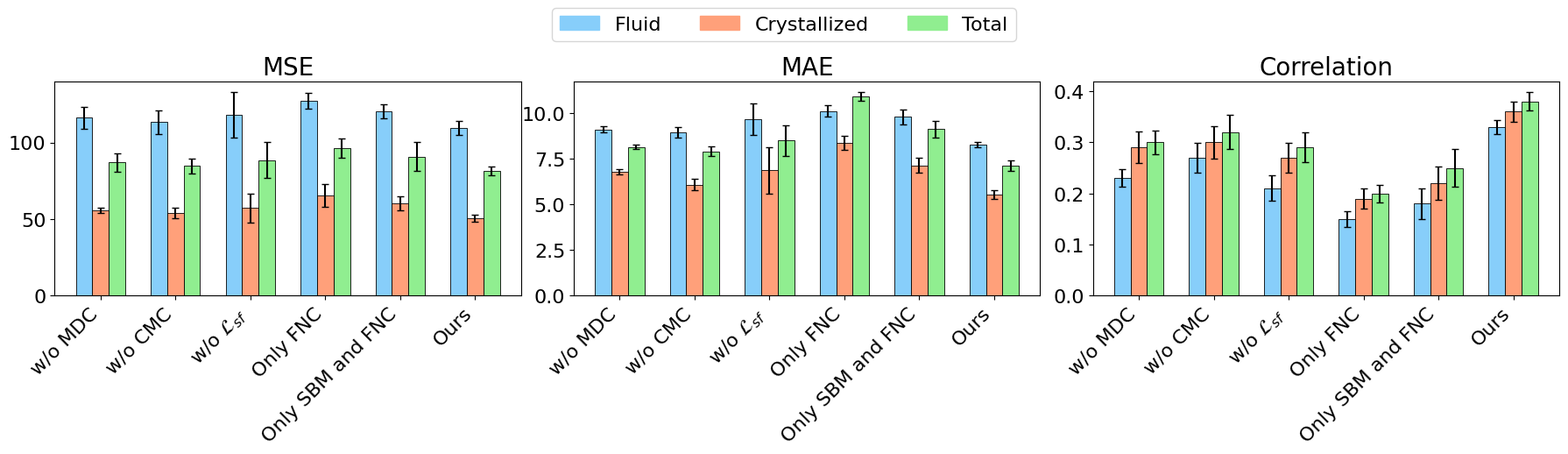}}
\caption{Outcomes of the performed ablation experiments (mean$\pm$standard deviation).} \label{fig2}
\vspace{-15 pt}
\end{figure}
\vspace{-5 pt}
\begin{figure}[t]
\centerline{\includegraphics[width=\columnwidth]{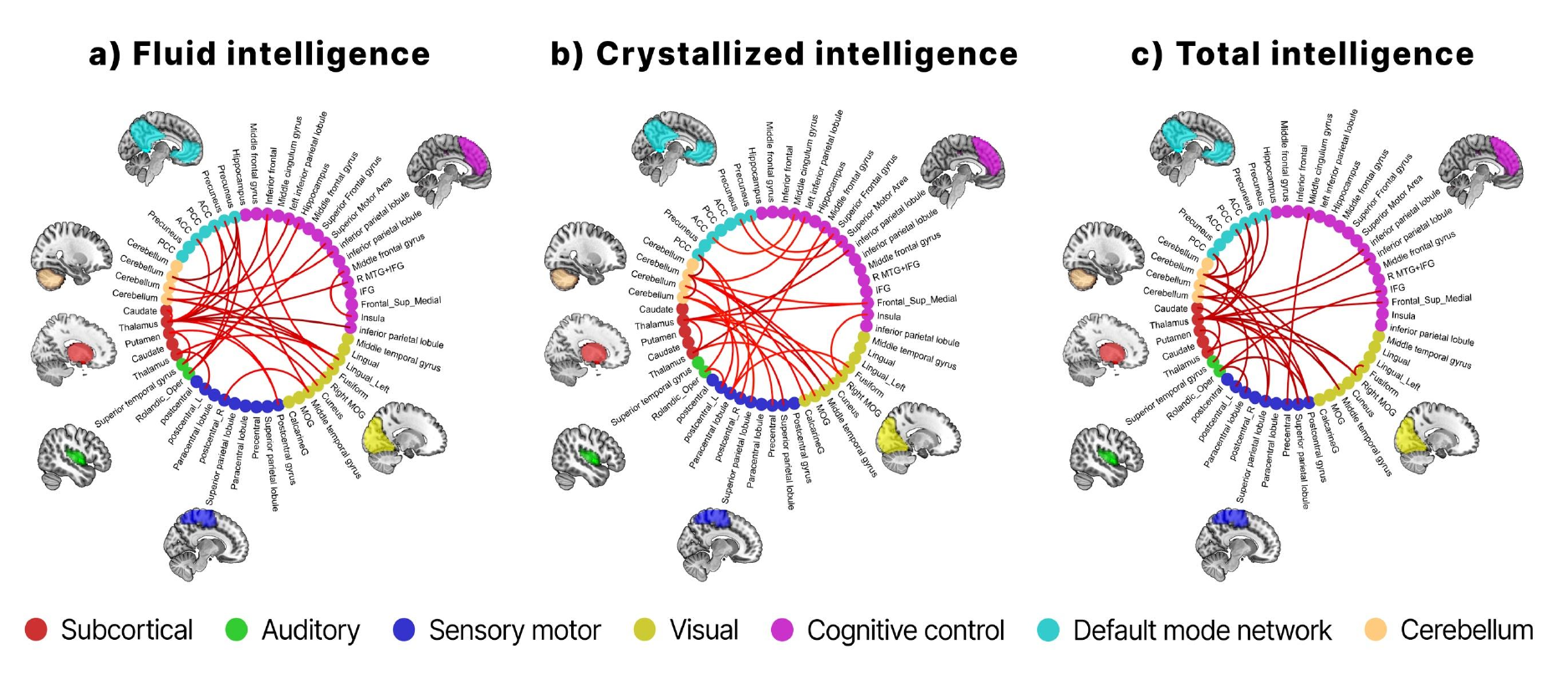}}
\caption{Top 3\% significant brain network connections identified for (a) fluid, (b) crystallized, and (c) total intelligence.} \label{fig3}
\vspace{-15 pt}
\end{figure}
\vspace{-10 pt}
\subsection{Comparison with State-Of-The-Art (SOTA) Methods}
\vspace{-5 pt}
We benchmarked our framework against SOTA GNN-based approaches including: GAT \cite{base1}, GT \cite{base2}, SFDN \cite{base3}, SFIN \cite{base4}, Joint GCN \cite{base5}, BrainNN \cite{base6}, GCNN \cite{base7}, and Joint DCCA \cite{base8}, in the intelligence prediction task. As reported in Table \ref{tab1}, our approach consistently outperformed all baselines, achieving the lowest MSE and MAE and the highest correlation scores across all three intelligence scores. Compared to SFDN, the strongest baseline, our model achieved a $0.04$ increase in fluid, $0.02$ in crystallized, and $0.03$ in total intelligence correlation, highlighting the impact of MDC over fixed radius DC. Although SFIN showed competitive correlation scores, it fell behind in MSE and MAE due to its reliance on direct structure-function interactions, lacking multi-level connectivity modeling and overlooking indirect pathways. Vanilla methods such as GAT and GT were limited by static attention mechanisms, whereas Joint GCN and GCNN failed to integrate MDC, affecting predictive accuracy. Similarly, BrainNN and Joint DCCA processed structural and functional modality separately, limiting proper multimodal alignment. 
\vspace{-10 pt}
\subsection{Ablation Experiments}
\vspace{-5 pt}
As demonstrated in Fig. \ref{fig2}, our ablation study was designed to evaluate the contribution of key components of our framework. Removing MDC (w/o MDC) resulted in a decline of $0.10$, $0.07$, and $0.08$ in correlation for fluid, crystallized, and total intelligence scores, respectively, highlighting the significance of indirect pathways in structure-function interplay. Excluding CMC (w/o CMC) led to a decrease of $0.06$ in correlation across all three intelligence scores, indicating their role in capturing multimodal interaction. Discarding structure-function consistency loss (w/o $\mathcal{L}_{sf}$) caused a huge increase in MSE and MAE, with correlation dropping by $0.12$ (fluid), $0.09$ (crystallized and total), demonstrating its impact in reinforcing structure-function alignment. Additionally, models trained on only FNC and only SBM+FNC showed inferior performance, demonstrating higher MSE, MAE and lower correlation, further confirming that MDC and CMC are essential for capturing cross-modal interaction in cognition.

\vspace{-13 pt}
\subsection{Interpretation of Significant Networks}
\vspace{-5 pt}
Fig. \ref{fig3} illustrates the top 3\% of structure-function connections identified by \textit{MAGNet}’s attention weights as most relevant to intelligence prediction, revealing distinct yet interconnected pathways critical for cognitive function. Fluid intelligence (Fig. \ref{fig3}a) was found predominantly linked to the subcortical (SC) and cognitive control (CC) networks, highlighting their role in executive function, adaptive reasoning and problem-solving \cite{exp_6}. Crystallized intelligence (Fig. \ref{fig3}b) was primarily associated with the CC and default mode (DM) networks, facilitating knowledge retrieval and semantic integration \cite{I_1}.
Total intelligence (Fig. \ref{fig3}c) exhibited concentrated but distributed connections involving key network across the CC, DM, visual, and sensorimotor networks, reflecting integrated processing across multiple systems relevant to overall cognitive function \cite{I_1}. The recurring involvement of the CC network across all intelligence measures signified its role in higher-order cognition \cite{exp_8}, validating our framework's capability to capture meaningful structure-function interactions.

\vspace{-15 pt}
\section{Conclusions}
\vspace{-10 pt}

We presented \textit{MAGNet}, a Transformer-based adaptive GNN framework that employs a hybrid brain graph to capture structure-function interplay and learn multimodal representations for prediction. Through multi-level connections and local-global attention with a joint loss, \textit{MAGNet} captures cross-modal interactions and provide cognitive insights. Experimental outcomes on the ABCD dataset confirmed its superiority over SOTA baselines. Future work will incorporate dynamic FNC, longitudinal analyses of neurobehavioral change, and broader neurocognitive and clinical applications.

\bibliographystyle{IEEEbib}
\bibliography{refs}

\end{document}